\documentclass[sigconf]{acmart}
\AtBeginDocument{%
  }

\copyrightyear{2026}
\acmYear{2026}
\setcopyright{cc}
\setcctype{by}
\acmConference[WWW '26]{Proceedings of the ACM Web Conference 2026}
{April 13--17, 2026}{Dubai, United Arab Emirates}
\acmBooktitle{Proceedings of the ACM Web Conference 2026 (WWW '26),
April 13--17, 2026, Dubai, United Arab Emirates}
\acmDOI{}
\acmISBN{}
\usepackage{url}
\usepackage{xspace}
\usepackage{enumitem}
\newcommand{\hide}[1]{}
\usepackage{multirow}
\usepackage{amsmath}
\usepackage[normalem]{ulem}
\useunder{\uline}{\ul}{}
\usepackage{mathrsfs}
\usepackage{subfigure}
\usepackage{color}
\usepackage{tabularx}
\usepackage{bbding} 
\usepackage{array}
\usepackage{graphicx}  
\usepackage{float}  
\usepackage{subfigure}  
\usepackage{makecell}
\usepackage{booktabs}
\usepackage{subcaption}
\usepackage{lipsum} 
\usepackage{placeins}
\usepackage{tcolorbox}
\newcommand{\acronym}[1]{\underline{\textbf{#1}}}
\newcommand{\data}{\textbf{TRACE}\xspace}
\newcommand{\oursys}{\textbf{READER}\xspace}
\definecolor{demphcolor}{RGB}{144, 144, 144}
\definecolor{mygray}{gray}{0.4}
\definecolor{lightgray}{rgb}{0.9, 0.9, 0.9}
\newcommand{\demph}[1]{\textcolor{demphcolor}{#1}}

\usepackage[dvipsnames]{xcolor}
\begin{document}

\title{Delayed Feedback Modeling for Post-Click Gross Merchandise Volume Prediction: Benchmark, Insights and Approaches}

\thanks{This paper has been accepted by the ACM Web Conference (WWW) 2026.
This is the camera-ready version.
Please refer to the published version for citation once available.}

\author{Xinyu Li}
\authornote{Equal Contribution.}
\email{xinyuli@stu.xmu.edu.cn}
\orcid{0009-0007-5575-8221}
\affiliation{
  \institution{School of Informatics, Xiamen University}
  \city{Xiamen}
  \country{China}
}

\author{Sishuo Chen}
\authornotemark[1]
\email{chensishuo.css@alibaba-inc.com}
\orcid{0009-0007-8845-5817}
\affiliation{
   \institution{Taobao \& Tmall Group of Alibaba}
   \city{Beijing}
   \country{China}
}

\author{Guipeng	Xv}
\email{xuguipeng@stu.xmu.edu.cn}
\orcid{0000-0001-5320-5489}
\affiliation{
  \institution{School of Informatics, Xiamen University}
  \city{Xiamen}
  \country{China}
}

\author{Li Zhang}
\email{zl428934@alibaba-inc.com}
\orcid{0009-0005-1311-4116}
\affiliation{
   \institution{Taobao \& Tmall Group of Alibaba}
   \city{Beijing}
   \country{China}
}

\author{Mingxuan Luo}
\email{luomingxuan@stu.xmu.edu.cn}
\orcid{0009-0006-7659-1320}
\affiliation{
  \institution{School of Informatics, Xiamen University}
  \city{Xiamen}
  \country{China}
}

\author{Zhangming Chan}
\email{zhangming.czm@alibaba-inc.com}
\orcid{0000-0002-3081-2427}
\affiliation{
   \institution{Taobao \& Tmall Group of Alibaba}
   \city{Beijing}
   \country{China}
}

\author{Xiang-Rong Sheng}
\email{xiangrong.sxr@alibaba-inc.com}
\orcid{0009-0006-4864-574X}
\affiliation{
   \institution{Taobao \& Tmall Group of Alibaba}
   \city{Beijing}
   \country{China}
}

\author{Han Zhu}
\email{zhuhan.zh@alibaba-inc.com}
\orcid{0000-0002-9522-5637}
\affiliation{
   \institution{Taobao \& Tmall Group of Alibaba}
   \city{Beijing}
   \country{China}
}

\author{Jian Xu}
\email{xiyu.xj@alibaba-inc.com}
\orcid{0000-0003-3111-1005}
\affiliation{
   \institution{Taobao \& Tmall Group of Alibaba}
   \city{Beijing}
   \country{China}
}

\author{Chen Lin}
\authornote{Corresponding author.}
\email{chenlin@xmu.edu.cn}
\orcid{0000-0002-2275-997X}
\affiliation{
  \institution{School of Informatics, Xiamen University}
  \city{Xiamen}
  \country{China}
}

\renewcommand{\shortauthors}{Xinyu Li et al.}

\begin{abstract}
The prediction objectives of online advertisement ranking models are evolving from probabilistic metrics like conversion rate (CVR) to numerical business metrics like post-click gross merchandise volume (GMV). Unlike the well-studied delayed feedback problem in CVR prediction, delayed feedback modeling for GMV prediction remains unexplored and poses greater challenges, as GMV is a continuous target, and a single click can lead to multiple purchases that cumulatively form the label. 

To bridge the research gap, we establish \data, a GMV prediction benchmark containing complete transaction sequences rising from each user click, which supports delayed feedback modeling in an online streaming manner.
Our analysis and exploratory experiments on \data reveal two key insights: 
(1) the rapid evolution of the GMV label distribution necessitates modeling delayed feedback under online streaming training;
(2) the label distribution of \textit{repurchase samples} substantially differs from that of single-purchase samples, highlighting the need for separate modeling.
Motivated by these findings, we propose \acronym{R}epurchas\acronym{E}-\acronym{A}ware \acronym{D}ual-branch pr\acronym{E}dicto\acronym{R} (\oursys), a novel GMV modeling paradigm that selectively activates expert parameters according to repurchase predictions produced by a router.
Moreover, \oursys dynamically calibrates the regression target to mitigate under-estimation caused by incomplete labels.
Experimental results show that \oursys yields superior performance on \data over baselines, achieving a 2.19\% improvement in terms of accuracy.
We believe that our study will open up a new avenue for studying online delayed feedback modeling for GMV prediction, and our \data benchmark with the gathered insights will facilitate future research and application in this promising direction.
Our code and dataset are available at \url{https://github.com/alimama-tech/OnlineGMV}.
\end{abstract}




\maketitle


\begin{figure}[th]
    \centering
    \includegraphics[width=0.95\columnwidth]{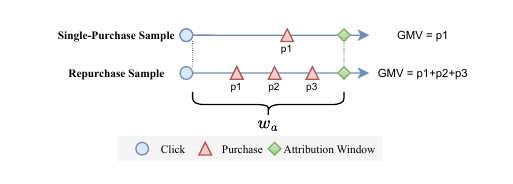}
    \caption{Label accumulation processes in GMV prediction, where the final label equals to the sum of the transaction prices of all purchases within the attribution window $w_a$.}
    \label{fig:intro}
\end{figure}

\section{Introduction}
As value-based bidding strategies such as \textit{Max Conversion Value} and \textit{Target ROAS (return on ad spend)}~\cite{zhu2017optimized,aggarwal2024auto,li2025beyond} play an increasingly significant role in web advertising, the prediction target of advertisement models has expanded from probabilistic metrics like click-through rate (CTR) and post-click conversion rate (CVR) to numerical business metrics such as \textit{post-click gross merchandise value} (\textbf{post-click GMV})~\cite{yu2025transun}, namely the total sales value attributed to an ad click conditioned on conversion. 

Considering that users may make \textbf{multiple repurchases}, the post-click GMV label of an ad click $c$ can be formulated as $\text{GMV}_{c} = \sum_{i=1}^{N} p_i$, where $N$ is the purchase count within the attribution window $w_a$, and $p_i$ is the transaction price of the $i$-th purchase.
Real-time bidding requires accurate predictions, yet ground-truth conversions--and thus the true GMV--only become available with a delay after a click occurs. To this end, modeling of this delayed feedback is therefore essential to maintain prediction accuracy.

As illustrated in Figure~\ref{fig:intro}, the GMV label associated with an ad click is updated up to $N$ times within the attribution window. This dynamic labeling process necessitates a delayed feedback modeling paradigm that effectively balances label accuracy and sample freshness~\cite{DFM,DEFUSE,Defer}. 
Although delayed feedback modeling for post-click CVR prediction has drawn sustained research attention~\cite{DFM,yoshikawa2018nonparametric,DEFUSE,Defer,ES-DFM,IF-DFM,KtenaTTMDHYS19,wang2020delayed,dai2023dually,liu2024online,yang2021capturing,chen2025see,chan2023capturing}, the challenge is overlooked in post-click GMV prediction.
Prior work on GMV prediction, such as TransSun~\cite{yu2025transun}, considers only single-purchase scenarios, ignoring the pronounced delayed feedback phenomenon in repurchase samples, which are common in real-world e-commerce scenarios.

To fill this gap, we construct \acronym{TRA}nsa\acronym{C}tion s\acronym{E}quences (\data), the first benchmark for post-click GMV prediction covering single-purchase and repurchase instances.
For each ad click leading to purchase(s), \data records the complete transaction sequence from the click until the end of the attribution window, so that the cumulative GMV trajectory is available for developing streaming models updated in an online manner.
\data is collected from the display advertising system of Taobao, Alibaba, and rigorously anonymized for public use,  providing a valuable testbed for developing delayed feedback models for GMV prediction.~\looseness=-1

To collect insights for algorithm development, we conduct data analysis and experiments on \data, making two key findings.
\textbf{First}, streaming models updated in an online manner remarkably surpass offline-trained counterparts, highlighting the significance of \textit{model freshness} and the value of delayed feedback modeling under \textit{online training}.
\textbf{Second}, the label distribution of single-purchase and repurchase samples significantly differs from each other, indicating that training a single model for both scenarios is challenging, and separate modeling  for each scenario may offer great potential.


Inspired by these insights, we propose \oursys, short for \acronym{R}epurchas\acronym{E}-\acronym{A}ware \acronym{D}ual-branch pr\acronym{E}dicto\acronym{R}, a novel online-updated GMV prediction model. \oursys features two specialized expert towers -- one tailored for single-purchase samples and the other for repurchase samples -- along with a lightweight router that dynamically selects the appropriate expert by predicting whether an ad click will lead to a single or multiple purchases. 
Beyond the architectural design, \oursys incorporates two tailored debiasing modules to better handle delayed feedback: 
(1) a label calibrator that maps the observed accumulated label to a pseudo-final label, mitigating underestimation caused by incomplete observations; 
(2) a partial label unlearning strategy applied when the attribution window closes to reduce residual bias.
Experiments show that \oursys significantly outperforms baselines on the \data benchmark, raising the accuracy  and AUC metrics by 2.19\% and 0.86\%, respectively.
Moreover, ablation experiments demonstrate the effectiveness of the key modules composing the \oursys model.

We believe that our study provides actionable insights for researchers and practitioners seeking to improve GMV prediction for web advertising, and we summarize our contributions as follows:

\begin{enumerate}[leftmargin=*]
    \item \textbf{Data Resources}. We release \data, the first benchmark for post-click GMV prediction with delayed feedback modeling, a significant yet ignored problem in web advertising.
    \item \textbf{Valuable Insights}. Our experiments reveal key findings, such as the importance of online training and dual-branch modeling for single-purchase versus repurchase samples.
    \item \textbf{Effective Approaches}. We propose \oursys, a novel framework for addressing the delayed feedback problem in post-click GMV prediction. Composed of a dual-branch architecture and meticulously designed optimization strategies for debiasing, \oursys achieves outstanding performance on \data.
\end{enumerate}

\section{Background and Related Works}

\subsection{GMV Prediction}

\subsubsection{Definition of GMV Targets}
The definition of GMV targets varies in the literature, depending on the specific application requirements, such as seller-level GMV~\cite{Gaia,GATGF}, user-level GMV~\cite{xin2019multi}, and order-level GMV~\cite{yu2025transun}.
To our knowledge, our study is the first to formally define and address the problem of post-click GMV prediction, which aims to forecast the total GMV attributable to an ad click that ultimately leads to conversions.
The most closely related problem setting to ours is the order-level GMV prediction~\cite{yu2025transun}, which focuses on estimating the monetary value of a single order but overlooks potential repurchases triggered by an ad click.

\subsubsection{The Significance of Post-Click GMV in Ad Ranking}
In popular value-based ad bidding modes such as \textit{Max Conversion Value} and \textit{Target ROAS (return on ad spend)}~\cite{zhu2017optimized,aggarwal2024auto,li2025beyond}, the Effective Cost Per Mille (ECPM) score of an online advertisement $a$ is formalized as
\begin{align} \label{eq:bid}
    \text{ECPM}(a) &= \text{pCTR}(a) \times \text{bid}(a) \nonumber \\
    &=  \text{pCTR}(a)\times \lambda \times \text{pCVR}(a)\times \text{pGMV}(a),
\end{align}
where $\text{pCTR}(a)$ is the predicted CTR, $\text{pCVR}(a)$ is the predicted CVR, $\text{pGMV}(a)$ is the predicted post-click GMV, $\lambda$ is the bidding parameter.
\textbf{Note}: although the price of the item in $a$ is available, simply replacing $\text{pGMV}(a)$ with the item price is sub-optimal, considering the existence of repurchases and the varying transaction price for each order due to consumers' selection of product variants and their use of coupons.
Therefore, accurate prediction of post-click GMV is essential for automated bidding and traffic allocation~\cite{aggarwal2024auto,li2025beyond}.

\subsection{Delayed Feedback Modeling}
Unlike the click feedback taking place immediately after ad exposure,
post-click targets like conversions (\textit{e.g.}, product purchase in e-commerce) are often substantially delayed relative to ad exposure, necessitating dedicated delayed feedback modeling techniques~\cite{DFM,zhao2023ecad}.
Generally, considering model freshness is important in highly dynamic online advertising scenarios, existing approaches usually devise \textbf{sample collection strategies}~\cite{ES-DFM,Defer,DEFUSE} for fresh label collection and online model updating before the end of the attribution window $w_a$.
Furthermore, given the gap between the fresh yet incomplete labels and the complete label available only after the end of $w_a$, previous studies have developed a wide array of \textbf{debiasing approaches}~\cite{DEFUSE,dai2023dually,liu2024online,yang2021capturing} to optimize the trade-off between model freshness and label correctness.
In spite of the significance and prevalence of delayed feedback modeling,
almost all of the existing studies focus on CVR prediction models that judge whether an ad click leads to conversion, overlooking the forecast of numerical business targets such as post-click GMV.
To the best of our knowledge, our work takes the first step to study delayed feedback modeling in GMV prediction.

\section{Benchmark and Insights}
\subsection{Benchmark Construction}
\subsubsection{Overview} 
As there is no publicly available dataset for post-click GMV prediction, we construct the \acronym{TRA}nsa\acronym{C}tion s\acronym{E}quences (\data) benchmark, which is sampled from the display advertising logs of Taobao, Alibaba.
The core characteristic of \data is the availability of the \textbf{complete purchase sequence} with timestamps and transaction prices for each ad click that leads to conversions.
This characteristic enables a systematic investigation of online-updated streaming models and delayed feedback modeling approaches—techniques shown to be beneficial for CVR prediction~\cite{DEFUSE,Defer} but largely overlooked in the context of GMV prediction.
The dataset covers sampled clicks in 82 days, and the attribution window is 7 days.
The statistics of \data are shown in Table~\ref{Tab:dataset}.~\looseness=-1

\subsubsection{Data Sampling Criteria}
Given the requirements for data confidentiality and the need for efficient academic experimentation, we sample ad clicks from industrial logs in two dimensions: advertising scenarios and users.
Regarding ad scenarios, we randomly sample 15 from the representative advertising scenarios on the Taobao app. For users, we apply a stratified sampling strategy -- assigning higher sampling rates to highly active users and lower rates to less active one -- to obtain a more representative user subset.
This dataset does not represent real-world business metrics or operational conditions.

\begin{table}[t] 
\caption{Statistics of \data, where repurchase samples denote the ad clicks bringing multiple purchases.}
\label{Tab:dataset}
\begin{tabular}{@{}cccccc@{}}
\toprule
Clicks     & Repurchase Samples  & Users     & Items     \\ \midrule
7.16m    & 3.84m (53.55\%)             & 3.60m & 1.96m \\ \bottomrule
\end{tabular}
\end{table}

\subsubsection{Feature and Label Schema}
Each ad click instance in \data contains four categories of signals:
\begin{itemize}
[leftmargin=10pt,topsep=2pt]
    \item \textit{Categorical features}: 22 user/item/context attributes;
    \item \textit{Temporal information}: The  click timestamp and the sequence of purchase timestamps;
    \item \textit{Original transaction prices}: The sequence of transaction prices aligned with the purchase timestamp sequence.
    \item \textit{Derived GMV labels}: The cumulative GMV trajectory and the final ground-truth GMV label.
\end{itemize}
\subsubsection{Data compliance and Availability}
The \data dataset undergoes strict de-identification and anonymization, where all user, item/product, campaign, and placement identifiers are irreversibly hashed, and any directly or indirectly identifying attributes are removed.
Only the fields essential for academic model development are retained, these do not reflect any real-world business scenarios due to the specific ad context and user sampling procedures employed. 
We have released the \data benchmark to pave the way for studies on post-click GMV prediction in the community.

\begin{table}[t]
\centering
\caption{Summary of Key Notations}
\label{Tab:notation-summary}
\resizebox{0.45\textwidth}{!}{
\begin{tabular}{ll}
\toprule
Symbol & Definition \\
\midrule
$x$ & Embedding vector of user/item/context features  \\
$t^c$
 & Timestamp of the click event \\

$\mathcal{P}=\left\{\left(t_{i}^{p}, p_{i}\right)\right\}_{i=1}^{N}$

 & Sequence of purchases (time and price) \\

$t_i^p$
 & Timestamp of the $i$-th purchase \\ 

$p_i$
 & Transaction price of the $i$-th purchase \\

$N$ & Total number of purchases \\

$y^*$
 & Final GMV label (target) \\

$y^{(t)}$
 & Observed GMV up to momment $t$ \\

$\tilde y^{(t)}$
 & Pseudo-label adjusted from $y^{(t)}$ \\

$\hat{y}$ & Predicted GMV \\
\bottomrule
\end{tabular}}

\end{table}

\subsection{Problem Setup}

We formalize each sample in the \data benchmark for post-click GMV prediction as  $c=⟨x,t,\mathcal{P}⟩$, where $x\in \mathbb{R}^d$ is the embedded categorical features, $t$ is the click timestamp, and $\mathcal{P}=\left\{\left(t_{i}^{p}, p_{i}\right)\right\}_{i=1}^{N}$ is the purchase sequence, in which  $t_i^p$ is the timestamp of the $i$-th purchase, $p_i$ is the transaction price of this purchase and $N$ is the number of purchases within the attribution window $w_a$.
The ground-truth label of the GMV prediction task is $y^{*}=\sum_{i=1}^{N} p_{i}$.
Considering $y^{*}$ is not available until the end of $w_a$, at a certain moment $t$, we can utilize the observed partial label ${y^{(t)}}=\sum_{t_i\leq t}p_i$ for model updating.
Table~\ref{Tab:notation-summary} summarizes the notations in the paper.

\begin{figure}[t]
    \centering
    \includegraphics[width=0.45\textwidth]{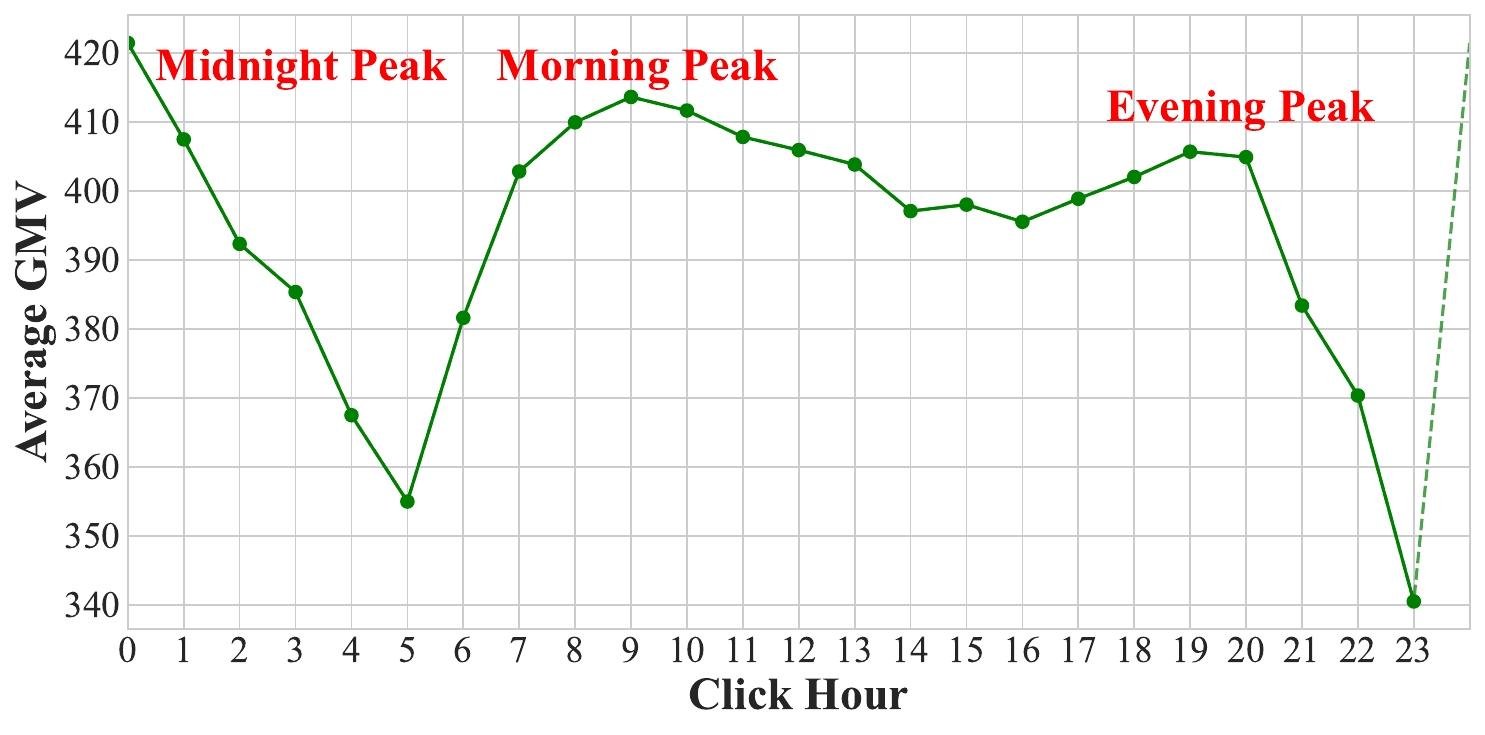}
    \caption{The hourly evolution of the average GMV label.}
    \label{fig:distribution_drift}
\end{figure}
\subsection{Insights from Analysis and Experiments} \label{subsec:insights}

Given the lack of public benchmarks or prior art in post-click GMV modeling, we conduct data analysis and exploratory experiments on the \data benchmark. 
Key insights are listed as follows:

\begin{tcolorbox}[colback=cyan!5!white, colframe=cyan!45!blue!60, title=\textbf{Takeaways}]
\begin{enumerate}[label=\arabic*., left=0pt] 
\item Rapid shifts in the GMV label distribution necessitate \textbf{delay feedback modeling under online streaming training}.
\item The label distribution of \textbf{single-purchase and repurchase samples} significantly differs from each other, which suggests that \textbf{repurchase-aware separate modeling} is beneficial.
\end{enumerate}
\end{tcolorbox}


\subsubsection{The Significance of Delayed Feed back Modeling in an Online-Training Manner.}
To capture the intra-day evolving pattern of GMV targets, we calculate the average GMV of ad clicks taking place in each hour and show its trend in Figure~\ref{fig:distribution_drift}.
We find significant hourly variations in the average GMV label,  with clear spikes during the midnight (00:00), morning (09:00), and evening (19:00) periods.
Such substantial fluctuations suggest that \textbf{streaming online training } is necessary to maintain model freshness and adapt to evolving user behavior.
To further verify the conjecture, we conduct exploratory experiments for comparing two model training regimes:~\looseness=-1
\begin{itemize} [leftmargin=10pt,topsep=2pt]
    \item \textbf{Offline Training}: The model only consumes the ground-truth label $y^{*}$ after the 7-day attribution window and keeps updated in a daily manner.
    \item \textbf{Online Training}: A vanilla streaming model that consumes the observed partial label $y^{(t)}$ when each purchase takes place.
\end{itemize}
As the AUC  metrics listed in Table~\ref{tab:odl}, the online-trained model significantly outperforms the daily-updated offline counterpart, which substantiates \textbf{the significance of model freshness and the superiority of online training}.

\begin{table}[t]
\centering
\caption{The performance advantage of online training over offline training in GMV prediction.}

\begin{tabular}{@{}c|c@{}}
\toprule
Training Regime           & AUC  \\ \midrule
Offline &  0.8055      \\
Online & 0.8165  \\ \bottomrule
\end{tabular} \label{tab:odl} 

\end{table}

\begin{figure}[t]
    \centering
    \includegraphics[width=0.4\textwidth]{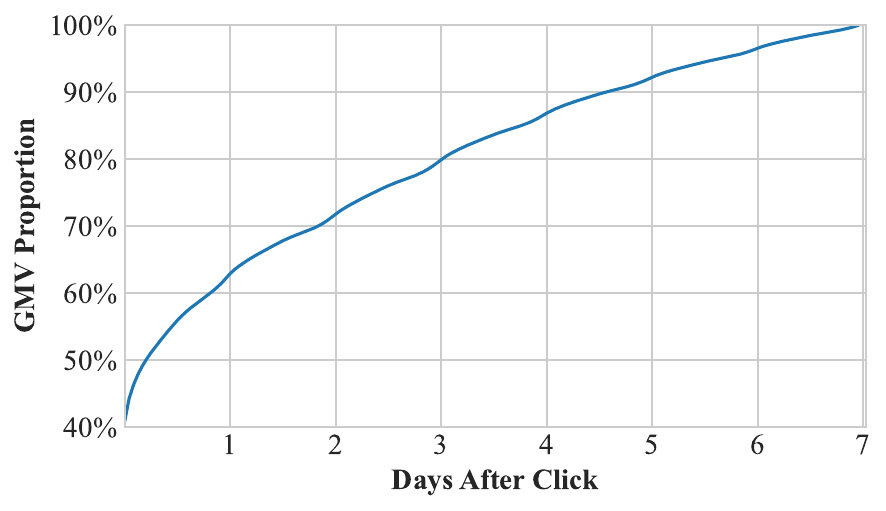}
    \caption{The cumulative proportion of GMV over time.}
    \label{fig:delayed_feedback}
\end{figure}

However, vanilla online learning on observed partial labels may suffer from label incorrectness and cause estimation bias in GMV prediction.
To verify the necessity of delayed feedback modeling for unbiased estimation, we plot the temporal evolution of the average cumulative fraction of realized GMV in \data in Figure~\ref{fig:delayed_feedback}.
We find that only 40\% of the final GMV comes from immediate purchases following ad clicks.
Consequently, the cumulative proportion of observed GMV rapidly increases to over 60\% after one day, then grows relatively steadily in the following days before the end of the attribution window, with the growth rate gradually slowing down.
As a substantial portion of the final GMV remains unobserved in the early post-click period, simply using partially observed GMV as the training target will lead to severe underestimation, and \textbf{modeling delayed feedback under online training} is essential for striking a balance between model freshness and label correctness.

\begin{figure}[t]
    \centering
    \includegraphics[width=0.4\textwidth]{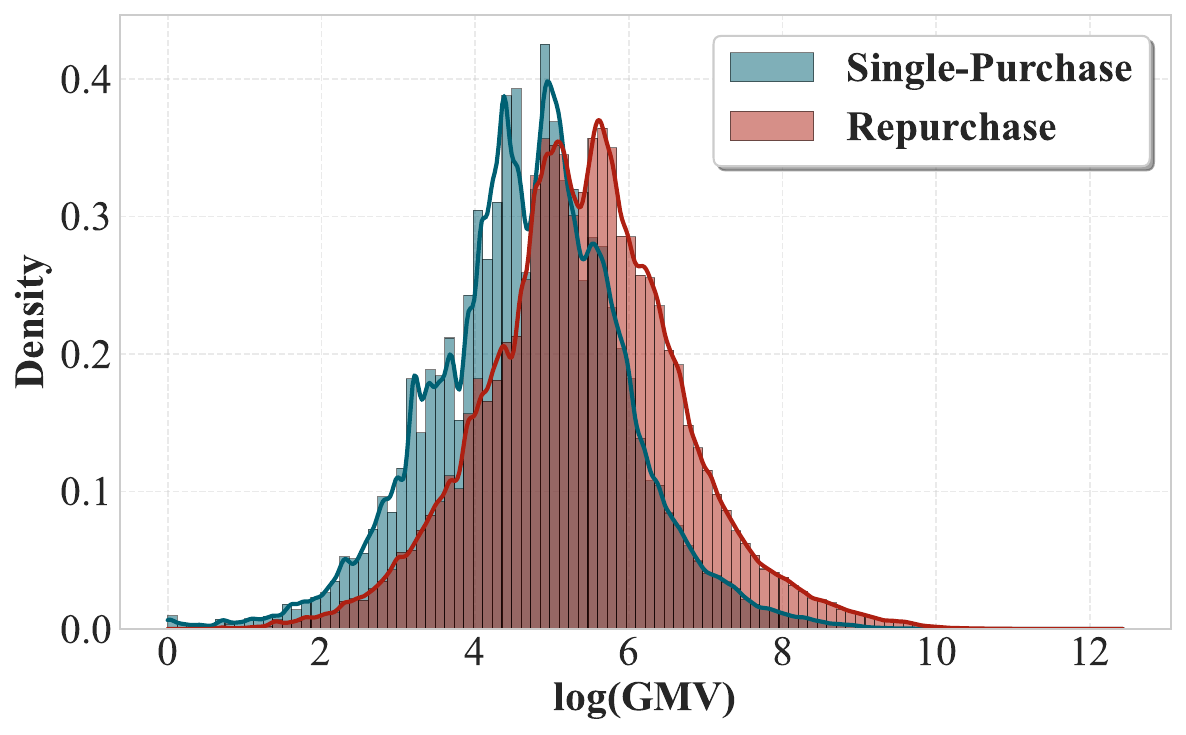}
    \caption{The GMV distributions of single-purchase and repurchase samples differ significantly.}
    \label{fig:dist_comp}
\end{figure}

\subsubsection{The Potential of Modeling Single-Purchase and Repurchase Samples Separately.}\label{two-mechanism}
The GMV distributions illustrated in Figure~\ref{fig:dist_comp}  reveal significant differences in label distribution between single-purchase and repurchase samples.
Concretely, repurchase samples exhibit a substantially higher average GMV and a more pronounced right-skewed tail compared to single-purchase ones.
Besides, we apply the two-sample Kolmogorov–Smirnov test and obtain a p-value at 0.00,  indicating strong statistical evidence of distributional discrepancy.
The distributional disparity suggests that treating these two types of samples within a unified model may pose challenges in model optimization, and \textbf{modeling single-purchase and repurchase instances separately} offers strong potential.

Such separate modeling necessitates a routing module that distinguishes between single-purchase and repurchase instances.
To assess the feasibility of such a router, we implement a lightweight MLP-based router, achieving an AUC of over 80\% on held-out test data.
This result indicates that repurchase prediction is a learnable task, and modeling single-purchase and repurchase samples in separate towers is therefore feasible.

\section{Methodology}

Inspired by the gathered insights in \S~\ref{subsec:insights}, we propose \acronym{R}epurchas\acronym{E}-\acronym{A}ware \acronym{D}ual-branch pr\acronym{E}dicto\acronym{R} (\oursys), a novel online-updated GMV prediction model under the repurchase-aware dual-branch architecture (introduced in \S~\ref{subsec:arch}) and enhanced by 
tailored debiasing strategies (introduced in \S~\ref{subsec:debias}) for delayed feedback modeling.
To effectively train the model, we design a two-stage training regime (introduced in \S~\ref{subsec:two-stage}).
Figure~\ref{fig:main} illustrates the workflow of \oursys.

 \begin{figure*}[t] \label{fig:main}
    \centering
    \includegraphics[width=0.75\textwidth]{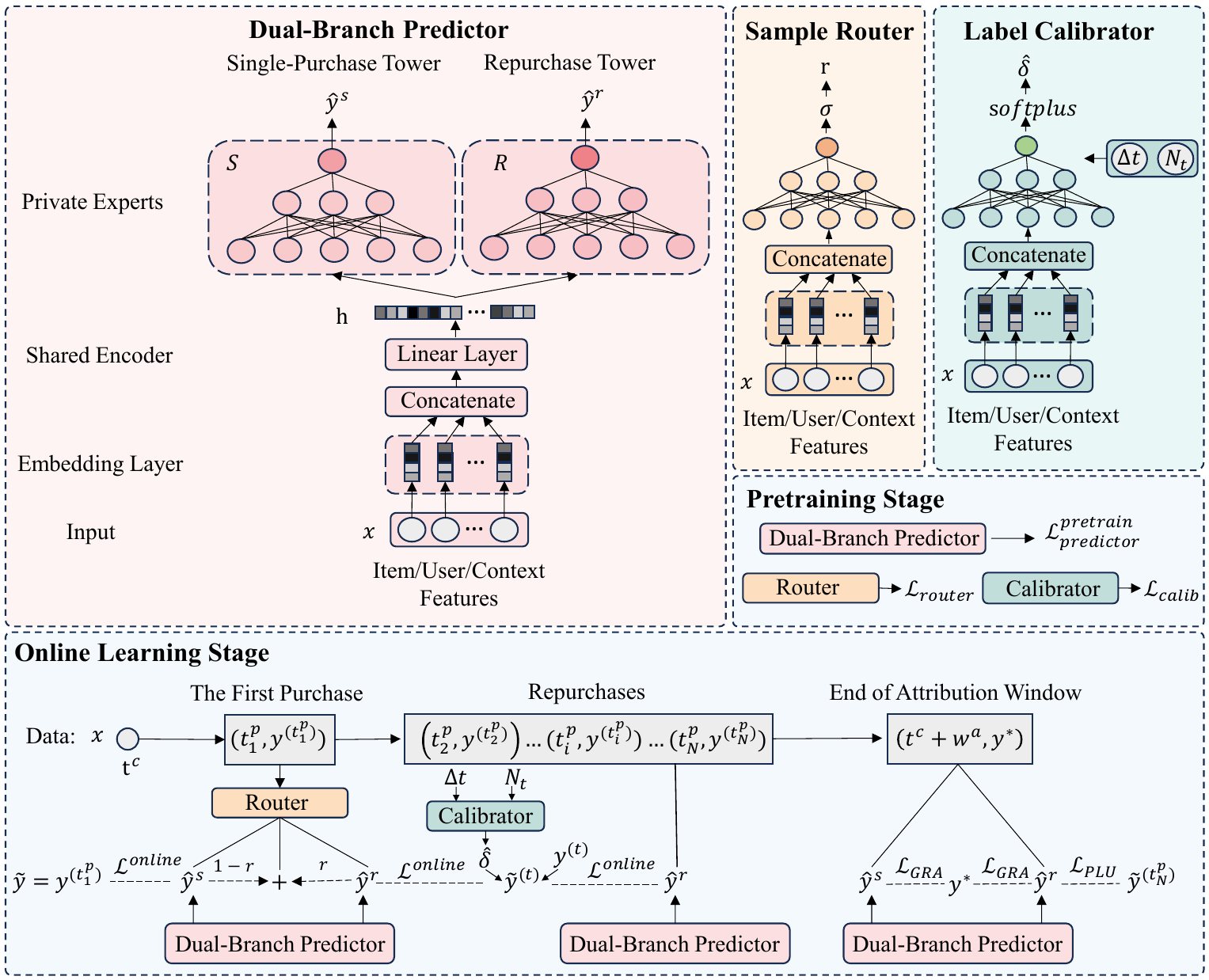}
    \caption{The workflow of our \oursys model.}
    \label{fig:main}
\end{figure*}

\subsection{Repurchase-Aware Dual-Branch Structure} \label{subsec:arch}

\subsubsection{Dual-Branch Predictor}
To address the distinct distributional characteristics between single-purchase and repurchase samples, we design a dual-branch architecture that decouples their prediction processes while enabling shared knowledge learning
We first use a shared encoder $f_{\theta}(\cdot)$ to get a feature vector $h$ from input features $x$:
\begin{equation}
    h =  f_\theta(\text{Emb}_t(x)),
\end{equation}
where $\text{Emb}_t(\cdot)$ is the shared embedding module, and $\theta$ denotes the parameters of the shared-bottom network.
To model the characteristics specific to each sample type, we place two separate MLP expert towers atop the shared bottom:~\looseness=-1
\begin{equation}
    \hat y^s = f_S(h), \\
    \hat y^r = f_R(h),
\end{equation}
where $f_S(\cdot)$ and $f_R(\cdot)$ represent the towers dedicated to single-purchase and repurchase samples, respectively. 

\subsubsection{Sample Router}
In online training under our dual-branch architecture, each sample must be assigned to the single-purchase or repurchase tower upon its first arrival, namely, when the first purchase takes place.
Considering joint learning the routing decision with GMV regression causes gradient entanglement and inter-tower interference, we equip \oursys with an independent sample router for making repurchase predictions.

Specifically, the sample router is implemented as a stand-alone network with its own embedding layer and MLP encoder, fully decoupled from the dual-branch predictor.
The function of the router $\phi$ can be formulated as follows:
\begin{equation}
    r=\sigma(f_\phi(\text{Emb}_{\phi}(x))/T),
\end{equation}
where $\text{Emb}_{\phi}(\cdot)$ and $f_{\phi}$ denote the  embedding and MLP, respectively, $T$ is the temperature parameter and $\sigma(\cdot)$ is the sigmoid function. The output $r\in(0,1)$ denotes the probability that the ad click leads to more than one purchase, which guides the routing as follows.~\looseness=-1

\subsubsection{Routing Mechanism} \label{soft-re}
When the single-purchase/repurchase label is unknown, such as in the inference stage and when the first purchase takes place in online training, \oursys relies on the router output $r$ for sample routing.
Concretely, the routing mechanism can be formalized as follows:
\begin{equation} \label{eq:routing}
\hat{y} = 
\begin{cases}
    \hat{y}^s , & \text{if } r \leq \tau_1 \quad \text{(single-purchase)} \\
    (1-r)\hat{y}^s + r\hat{y}^r & \text{if } \tau_1 < r < \tau_2 \quad \text{(hybrid)} \\
    \hat{y}^r & \text{if } r \geq \tau_2 \quad \text{(repurchase)}
\end{cases}
\end{equation}
where $\hat{y}$ is the predicted GMV label, $\hat{y}^s$ and $\hat{y}^r$ are the predictions produced by the single-purchase and repurchase towers, respectively, and $\tau_1,\tau_2$ are thresholds set to 0.1 and 0.9, respectively.
This three-zone routing mechanism handles uncertainty gracefully: low- and high-confidence samples are routed to specialized towers, while ambiguous cases benefit from a weighted combination of both.
The soft interpolation in the hybrid zone enables differentiable training and mitigates the risk of erroneous hard assignments.

\subsection{Debiasing Strategies} \label{subsec:debias}

To alleviate the bias introduced by the observed partial labels in online training, we devise two debiasing approaches---the \textbf{label calibrator (LC)} and \textbf{partial label unlearning (PLU)} on top of ground-truth alignment (GRA).

\subsubsection{Label Calibrator} In online training, the observed cumulative GMV $y^{(t)}$ at a moment $t$ is an underestimate of the final GMV $y^*$ in repurchase samples. 
If such underestimated values are directly used as regression targets, the model will suffer from biased supervision.
To mitigate the bias, we introduce a calibrator for adjusting the learning target of the repurchase tower.
The calibrator explicitly learns the discrepancy between the partial cumulative GMV $y^{(t)}$ and the final GMV $y^*$.
Specifically, the calibrator $\psi$ takes as input the feature embedding of $x$, the elapsed time $\Delta t$ since the click, and the number of purchases observed so far $N_t$:
\begin{equation} \label{eq:pred_gap}
    \hat\delta = \text{softplus}(f_{\psi}(\text{Emb}_{\psi}(x), \Delta t, N_t))
\end{equation}
where $\text{Emb}_{\psi}(\cdot)$ and $f_{\psi}(\cdot)$ are the calibrator-specific embedding and MLP. The elapsed time is formulated as:
\begin{equation}
\Delta t = \frac {t_i^p-t^c}{w_a},
\end{equation}
where $t_i^p$ denotes the purchase time, $t^c$ denotes the click time, and $w_a$ denotes the attribution window length.
The output $\hat \delta$ is the predicted difference between the log-transformed partial label and ground-truth label, which is an approximation of the true gap: 
\begin{equation} \label{eq:real_gap}
    \delta = \text{log}(1+y^*) - \text{log}(1+y^{(t)}).
\end{equation}
Then the calibrated label
$
    \tilde y^{(t)} = \text{exp}(\text{log}(1+y^{(t)})+\hat \delta) - 1
    \label{eq:pseudo-label}
$
will be closer to the ground-truth GMV $y^*$. 
The label calibrator provides more accurate supervision in online training, mitigating bias and improving training stability.

\subsubsection{Partial Label Unlearning on Top of Ground-Truth Alignment}
In online training, the calibrated labels produced by the calibrator provide timely supervision before the attribution window closes. 
However, when the attribution window ends, discrepancies between the calibrated pseudo $\tilde y$ and the ground-truth label $y^*$ will remain. 
To mitigate the bias, we first perform \textbf{ground-truth alignment (GRA)}, namely fitting the model with ground-truth labels.
Specifically, each sample is routed to the single-purchase or repurchase tower according to its purchase times, and the model is updated with the complete label $y^*$ for minimizing the target:
\begin{equation} \label{eq:gra}
    \mathcal{L}_\text{GRA}=\frac{1}{|\mathcal{D_\text{GR}}|}\sum_{x\in \mathcal{D}_{\text{GR}}}|\log(1+\hat y_x)-\log(1+y^*_x)|,
\end{equation}
where  $\mathcal{D}_\text{GR}$ denotes the set of samples with ground-truth labels.

On top of GRA, we propose \textbf{partial label unlearning (PLU)} to further mitigate the bias caused by the gap between the ground-truth GMV $y^*$ and the calibrated label $\tilde y^{(t_N^p)}$, where $t_N^p$ indicates the moment when the last purchase takes place.
Note that the calibrator is designed to predict the gap between the partial label $y^{(t)}$ and the final GMV $y^*$, so the corrected pseudo-label $\tilde y^{(t_N^p)}$ for the last observed purchase will incorrectly inflate the ground-truth label.
To mitigate this negative effect while preserving useful patterns learned from earlier updates, we hence introduce a partial label unlearning strategy, which is inspired by machine unlearning~\cite{bourtoule2021machine}.
Specifically, when the complete label $y^*$ becomes available, we maximize the following target for  counteracting the label inflation:
\begin{equation} \label{eq:plu}
    \mathcal{L}_\text{PLU}=\frac{1}{|\mathcal{D_\text{GR}}|}\sum_{x\in \mathcal{D}_{\text{GR}}}|\log(1+\hat y_x)-\log(1+\tilde y^{(t_N^p)}_x)|.
\end{equation}
PLU prevents the label inflation in the last purchase from misleading the training process, which reduces residual bias and stabilizes model training together with GRL.

\subsection{Two-Stage Training Regime} \label{subsec:two-stage}

For model warmup on ground-truth labels, we devise a two-stage training regime, which first pretrains the model in an offline manner on historical data with complete labels, and then conducts online learning in the real-time data stream.

\subsubsection{Pretraining Stage}
In the pretraining stage, the complete label $y^*$ and the total number of purchases $N$ is available for each instance, which we leverage to initialize all components of the model.
Regarding the \textbf{dual-branch predictor}, we perform sample routing according to the purchase count $N$ to get the prediction $\hat y$:
\begin{equation} \small
\hat{y} = 
\begin{cases}
    \hat{y}^s , & \text{if } N = 1 \quad \text{(single-purchase)} \\
    \hat{y}^r & \text{if } N > 1 \quad \text{(repurchase)}
\end{cases}
\end{equation}
Then we minimize the log-MAE loss~\footnote{We adopt the LogMAE loss due to its superiority over other popular choices (\textit{e.g.}, MAE, MSE, and log-MSE) on long-tailed GMV targets in our exploratory experiments.} for fitting the ground-truth GMV label $y^*$ on the pretraining dataset $\mathcal{D_\text{PRE}}$:
\begin{equation} \small
\mathcal{L}_{\text{predictor}}^{\text{pretrain}}=\frac{1}{|\mathcal{D_\text{PRE}}|}\sum_{x \in \mathcal{D_\text{PRE}}}^M\left|\log(1+\hat{y}_x)-\log(1+y^*_x)\right|,
\end{equation}

As for the \textbf{sample router}, the ground-truth label is $y^*_\text{router} = \mathbb{I}\{N>1\}$,
where $y^*_{\text{router}}=0$ for single-purchase samples and $y^*_{\text{router}}=1$ for repurchase ones.
We minimize the cross-entropy loss:~\looseness=-1
\begin{equation} 
\resizebox{0.42\textwidth}{!}{$
\mathcal{L}_{\text{router}}=-\frac{1}{|\mathcal{D_\text{PRE}}|}\sum_{x\in \mathcal{D_\text{PRE}}}\left[y^*_{\text{router}}\log r+(1-y^*_{\text{router}})\log(1-r)\right].$
}
\end{equation}

Regarding the \textbf{label calibrator}, since it aims to predict the gap between the observed partial GMV and the final GMV, it is trained on the subset $\mathcal{D_\text{RE}}$ composed of the repurchase samples in $\mathcal{D_\text{PRE}}$.
We minimize the absolute difference between the predicted label gap $\hat\delta$ and the true gap $\delta$ defined in \eqref{eq:pred_gap} and \eqref{eq:real_gap}:
\begin{equation} 
    \mathcal{L}_{\text{calib}}=\frac{1}{|\mathcal{D}_\text{RE}|}\sum_{x \in \mathcal{D}_\text{RE}}|\hat \delta_x - \delta_x|.
\end{equation}

\subsubsection{Online Learning Stage}
In the online learning stage, the model must learn from the partial labels that arrive sequentially as the user makes purchases.
When the first purchase takes place, the sample type is predicted by the sample router, and the predicted GMV $\hat y$, namely the output of the dual-branch predictor, is calculated following the routing rule defined in \eqref{eq:routing}.
Accordingly, the calibrated target $\tilde y^{(t)}$ for the predictor is defined as:
\begin{equation}
    \tilde y^{(t)} = (1-r)\cdot y^{(t)} + r \cdot (\text{exp}(\text{log}(1+y^{(t)})+\hat\delta) - 1),
\end{equation}
where $y^{(t)}$ is the observed partial label, $p$ is the predicted repurchase probability, and $\hat\delta$ is the the correction term predicted by the calibrator~\footnote{Although online training is also feasible for the label calibrator, we freeze it in the online learning stage as we find online updating only yields marginal gains.} as defined in \eqref{eq:pred_gap}.
Given the predicted GMV $\hat y$ and the calibrated label $\tilde y$, we update the parameters of the dual-branch predictor and the router to minimize the log-MAE loss:
\begin{equation}
    \mathcal{L}^{\text{online}}=\frac{1}{|\mathcal{D_\text{ON}}|}\sum_{x\in\mathcal{D_\text{ON}}}\left|\log(1+\hat{y}_x)-\log(1+\tilde y_x)\right|,
\end{equation}
where $\mathcal{D_\text{ON}}$ denotes the online data stream.
The overall optimization target  is to minimize the weighted sum of $\mathcal{L}^{\text{online}}$ and the debiasing targets $\mathcal{L}_\text{GRA}$ and $\mathcal{L}_\text{GRA}$ defined in \eqref{eq:gra} and \eqref{eq:plu}, respectively:
\begin{equation}
    \mathcal{L}^{\text{overall}} = \mathcal{L}^{\text{online}} + \lambda_1(\mathcal{L}_\text{GRA} - \lambda_2\mathcal{L}_\text{PLU}),
\end{equation}
where $\lambda_1$ and $\lambda_2$ are hyperparameters set to 0.1 and 0.5, respectively.


\section{Experiment}
\subsection{Experimental Setting}
 
\subsubsection{Baselines}

We compare our \oursys model with the following baselines under the vanilla single-tower architecture:
\begin{itemize}
[leftmargin=10pt,topsep=2pt]
\item  \textbf{Pre-Single}: The offline-trained model only consumes the pretraining data and then keeps frozen.
\item \textbf{Offline-Single}: The model only consumes the ground-truth label $y^*$ and keeps updated once a day.
\item  \textbf{Online-Single}: The model consumes the observed partial label $y^{(t)}$ evevy time a purchase takes places.
\end{itemize}
We also test the variants of Pre-Single and Offline-Single under our dual-branch structure, \textit{i.e.}, \textbf{Pre-Dual}, \textbf{Offline-Dual}, and \textbf{Online-Dual}.
Besides, we test oracle models \textbf{Oracle-Single} and \textbf{Oracle-Dual} under the single-branch and dual-branch architecture, respectively, which are trained on the ground-truth label when the first purchase takes place, which contradict the setting of real-world applications but provide performance upper bounds for reference.

\subsubsection{Training Setting}
All models except Pre-Single and Pre-Dual are trained under the two-stage regime for fair comparison.
The pretraining is conducted on the ad clicks in \data taking place from the 0-th  to the 50-th day, while the online learning is performed from the 57-th  to the 82-nd day.
All models are trained for one epoch~\cite{zhang2022towards}~\footnote{Altough a repurchase sample is consumed more than once in online training, we do not observe one-epoch over-fitting as the label keeps changing~\cite{DEFUSE}.} and the learning rate is searched from \{1\text{e}-2, 1\text{e}-3, 1\text{e}-4\}.~\looseness=-1

\subsubsection{Evaluation Metrics}
We adopt the following three evaluation metrics, whose details are given in Appendix~\ref{app}. 
\begin{itemize}
[leftmargin=10pt,topsep=2pt]
  \item \textbf{Area Under the ROC Curve (AUC)} widely used in ranking;
  \item \textbf{Accuracy (ACC)}: The proportion of samples for which the relative error between the predicted and true GMV  is within 20\%;
\item  \textbf{Absolute Predition Error (ALPR)}: The average of the logarithmic relative errors.
\end{itemize}

\subsection{Main Results}

\begin{table}[t]
\centering
\caption{GMV prediction performance. $\uparrow$ ($\downarrow$) indicates higher (lower) is better. The best results are highlighted in \textbf{bold}, the best baselines are underlined, and the metrics of the oracle models for reference are \demph{grayed out}. The last row gives the relative gain achieved by \oursys over the best baseline.~\looseness=-1}
\begin{tabular}{@{}l|ccc@{}}
\toprule
Models                             & AUC$\uparrow$                         & ACC$\uparrow$         & ALPR$\downarrow$                       \\ \midrule
\demph{Oracle-Single} & \demph{0.8168} & \demph{0.2580} & \demph{0.7823} \\ 
\demph{Oracle-Dual}                            &  \demph{0.8486}          &  \demph{0.4134}            & \demph{0.6273}            \\ \midrule
\emph{Baselines} \\
Pre-Single &0.8035 &0.2488 &0.8220 \\
Offline-Single &0.8055 & \underline{0.2556} &0.8125 \\
Online-Single & \underline{0.8165}  &0.2495 & \underline{0.8079} \\ \midrule
\emph{Our Approaches} \\
Pre-Dual                    & 0.8045                      & 0.2466                     & 0.8213                     \\
Offline-Dual                    & 0.8114                      & 0.2507                     & 0.7974                     \\
Online-Dual                    & 0.8161                      & 0.2562                     & 0.8506                     \\
\oursys                 &  \textbf{0.8235}               & \textbf{0.2612}             & \textbf{0.7523}               \\
Performance Gain & \textbf{\textcolor{OliveGreen}{+0.86\%}} & \textbf{\textcolor{OliveGreen}{+2.19\%}} & \textbf{\textcolor{OliveGreen}{-6.88\%}} \\ 
\bottomrule
\end{tabular}
\label{Tab:main}
\end{table}

We display the primary results in Table~\ref{Tab:main} and make two key observations as follows.

\textit{First, our \oursys model outperforms the baselines in terms of all metrics.}
Regarding AUC and ALPR, \oursys relatively raises AUC by 0.86\% and reduces ALPR by 6.88\% compared with the strongest baseline Online-Single;
as for the ACC metric, \oursys surpasses the best baseline Offline-Single by 2.19\% relatively.
Notably, \oursys outperforms Oracle Single and achieves performance closest to Oracle-Dual, showing the superiority of \oursys and the potential of repurchase-aware dual-branch modeling.~\looseness=-1

\textit{Second, the repurchase-aware dual-branch structure yields significant performance gains.}
A comparison between the single-branch models (Oracle-Single, Pre-Single, and Offline-Single) and their dual-branch counterparts reveals that the proposed repurchase-aware dual-branch architecture consistently yields performance improvements. 

We notice that in the online training setting (Online-Single \textit{v.s.} Online-Dual), the dual-branch structure leads to an improvement in ACC but deterioration of AUC and ALPR.
This suggests that the dual-branch structure needs to be enhanced with the debiasing strategies as done in \oursys in online training, for which we conduct comprehensive ablation experiments as follows.


\subsection{Ablation Study}

\begin{table}[t]
\centering
\caption{The model performance when ablating debiasing approaches. Calib denotes the label calibrator, GRA denotes the ground-truth alignment approach, and PLU denotes the partial label unlearning strategy.}
\begin{tabular}{@{}ccc|ccc@{}}
\toprule
 Calib & GRA & PLU  & AUC$\uparrow$     & ACC$\uparrow$     & ALPR$\downarrow$    \\ \midrule
 \XSolidBrush &\XSolidBrush &\XSolidBrush & 0.8161 & 0.2562 & 0.8506 \\
\Checkmark &\XSolidBrush &\XSolidBrush & 0.8180 & 0.2568 & 0.7738 \\
\Checkmark & \Checkmark &\XSolidBrush   & \textbf{0.8235} & 0.2604 & 0.7534 \\
   \Checkmark & \Checkmark & \Checkmark    & \textbf{0.8235} & \textbf{0.2612} & \textbf{0.7523} \\ \bottomrule
\end{tabular}
\label{tab:debias_ablation}
\end{table}

\subsubsection{Ablation on Debiasing Approaches}
To verify the effectiveness of our debiasing approaches introduced in \S~\ref{subsec:debias}, we ablate the choice of debiasing methods and present the results in Table~\ref{tab:debias_ablation}.
We find that \textbf{each debiasing method contributes to the performance improvement}.
Specifically, using only the label calibrator (Calib) yields a modest improvement in AUC alongside a substantial reduction in ALPR, highlighting its effectiveness in mitigating bias.
Adding ground-truth alignment (GRA) further boosts AUC by +0.0055 and improves ACC, highlighting the value of the delayed ground-truth label coming after the end of the attribution window.
Moreover, further incorporating partial label unlearning (PLU) leads to the best ACC and lowest ALPR, with no loss in AUC, demonstrating the complementary nature of the three debiasing strategies.~\looseness=-1

\subsubsection{Ablation on Architecture Design}
We ablate the model structure under the oracle setting where the final GMV label $y^*$ comes at the moment when the first purchase takes place.
This oracle setting eliminates noise caused by delayed feedback and errors from incorrect routing decisions, thereby providing a clean upper bound on the performance achievable by each architecture.
Although this setup is not feasible for real-time online training, it serves as an idealized diagnostic to assess capability of different architectural designs.
We compare four architectures:
\begin{itemize}
[leftmargin=10pt,topsep=2pt]
\item \textbf{Single-Branch}: Sharing one predictor for all samples.
\item \textbf{Independent Dual-Branch}: Single-purchase and repurchase cases are trained on completely separate towers.
\item \textbf{Frozen-Bottom Dual-Branch}: Sharing a fixed feature extractor but uses separate heads.
\item \textbf{Shared-Bottom Dual-Branch}: Both towers share a trainable bottom encoder but have independent heads.
\end{itemize}
As shown in Table~\ref{Tab:architecture}, \textbf{the Shared-Bottom Dual-Branch variant achieves the best performance}, outperforming the frozen-bottom variant and substantially surpassing the independent dual-branch and single-branch baselines.
These results sho that: (1) sharing low-level representations effectively captures common behavioral cues across both sample types, (2) enabling the bottom encoder to adapt online is crucial for performance, and (3) modeling heterogeneous distributions within a single tower is suboptimal—findings that align with the analysis in \S~\ref{two-mechanism}. Collectively, they underscore the superiority and necessity of our architectural design.~\looseness=-1
\begin{table}[t]
\centering
\caption{Architecture comparison under the oracle setting.}
\begin{tabular}{@{}l|ccc@{}}
\toprule
Architecture                                     & AUC$\uparrow$                              & ACC$\uparrow$                              & ALPR$\downarrow$           \\ \midrule
Single-Branch                & 0.8168          & 0.2580          & 0.7823 \\
Independent Dual-Branch            & 0.8348          & 0.3772          & 0.6819      \\
Frozen-Bottom Dual-Branch & 0.8404          & 0.3902         & 0.6593  \\       
Shared-Bottom Dual-Branch       & \textbf{0.8520} & \textbf{0.4124} & \textbf{0.6199} \\ \bottomrule
\end{tabular}
\label{Tab:architecture}
\end{table}

\subsubsection{Ablation on Routing Strategies}

Table~\ref{Tab:routing strategy} compares the repurchase-aware dual-branch predictor under two routing strategies:
\begin{itemize} [leftmargin=10pt,topsep=2pt]
\item \textbf{Hard Routing}: Choosing the single-purchase branch when $r<0.5$ or the repurchase branch when $r\geq0.5$.
\item \textbf{Hybrid Routing}: The three-zone routing strategy introduced in \eqref{eq:routing} in \S~\ref{soft-re}, which feeds low- and high-confidence samples to specialized branches, while interpolating the predictions of two branches for ambiguous cases.
\end{itemize}
We find that \textbf{the hybrid routing strategy adopted in \oursys beats the hard routing baseline} in terms of all metrics, which demonstrates the effectiveness of our proposal.
This highlights the importance of routing design in the repurchase-aware dual-branch predictor, and developing more advanced routing strategies is a promising direction for future improvement.
\begin{table}[t]
\centering
\caption{The performance under different routing strategies.}
\begin{tabular}{@{}l|ccc@{}}
\toprule
\multicolumn{1}{l}{}                  & AUC$\uparrow$     & ACC$\uparrow$     & ALPR$\downarrow$    \\ \midrule
Hard Routing & 0.8066 & 0.2538 & 0.8566 \\
Hybrid Routing      & 0.8161 & 0.2562 & 0.8506 \\ \bottomrule
\end{tabular}
\label{Tab:routing strategy}
\end{table}

\section{Conclusion}

In this work, we address a critical yet overlooked challenge in online advertising: delayed feedback modeling for post-click GMV prediction. Aware of the lack of data resources and prior art in the problem, we construct \data, the first publicly available post-click GMV prediction benchmark covering the entire purchase event sequence arising from each ad click, which lays the foundation for studying online training regimes and delayed feedback modeling.
Inspired by the insights collected on \data, such as the advantage of online streaming training and the potential of modeling single-purchase and repurchase instances separately, we present \oursys, a novel GMV modeling paradigm composed of the repurchase-aware dual-branch structure and tailed debiasing strategies for online learning.
Extensive experimental results on \data demonstrate the superiority of \oursys over strong baselines, and ablation results verify the effectiveness of the core modules composing our \oursys model.
We believe that our benchmark, analysis and approaches will offer practical insights for researchers and practitioners aiming to enhance GMV prediction in web advertising.

\begin{acks}
This work was supported by the Natural Science Foundation of China (No.62432011, 62372390), Science Fund for Distinguished Young Scholars of Fujian Province (No. 2025J010001), and Alibaba Group through Alibaba Innovative Research Program. 
Chen Lin is the corresponding author. This work was completed while the first author was an intern at Alibaba. 
\end{acks}

\clearpage

\bibliographystyle{ACM-Reference-Format}
\bibliography{sample-base}

\appendix
\appendix

\section{Evaluaion Metrics} \label{app}

Assuming $y$ is the label and $\hat y$ is the prediction, the evaluation metrics are defined as follows:
\begin{itemize}
[leftmargin=10pt,topsep=2pt]
  \item \textbf{Area Under the ROC Curve (AUC)}: 
In regression problems, the AUC is defined as the probability that, for a randomly selected pair of samples $(i, j)$, the model's predictions reflect the true ranking of the targets:
\begin{equation}
    \text{AUROC} = \mathbb{P}\big((\hat{y}_i - \hat{y}_j)(y_i - y_j) > 0\big)
\end{equation}
 This measures the proportion of correctly ordered pairs in terms of rank agreement. An AUC of 1 indicates perfect ranking and 0.5 corresponds to random performance.
  \item \textbf{Accuracy (ACC)}: The proportion of samples for which the relative error between the predicted and true GMV  is within 20\%:
  \begin{equation}
      \text{ACC} = \frac{1}{n} \sum_{i=1}^{N} \mathbb{I}\left( \frac{| \hat{y}_i - y_i |}{|y_i|} \leq 0.2 \right),
  \end{equation}
  where $N$ is the sample count, and $\mathbb{I}(\cdot)$ is the indicator function that equals 1 if the condition is satisfied and 0 otherwise.
\item  \textbf{Absolute Predition Error (ALPR)}: The average of the logarithmic relative errors:
\begin{equation}
    \text{ALPR} = \frac{1}{n} \sum_{i=1}^{N} \left| \log_2 \left( \frac{\hat{y}_i}{y_i} \right) \right|,
\end{equation}
where $N$ is the sample count. This metric measures the average multiplicative deviation between predictions and true values in log scale.
Compared to traditional metrics such as mean squared error (MSE) and mean absolute error (MAE), ALPR is less sensitive to extreme values with large targets and better reflects overall predictive accuracy across different scales. 
\end{itemize}

\end{document}